\documentclass{article}

\usepackage{arxiv}

\usepackage[utf8]{inputenc}
\usepackage[T1]{fontenc}
\usepackage{amsmath,amsfonts}
\usepackage{graphicx}
\usepackage{booktabs}
\usepackage{nicefrac}
\usepackage{microtype}
\usepackage{array}
\usepackage{float}
\newcolumntype{P}[1]{>{\raggedright\arraybackslash}p{#1}}

\usepackage{tikz}
\usetikzlibrary{shapes.geometric, arrows.meta, positioning}

\usepackage{pgfplots}
\pgfplotsset{compat=1.16} 
\usepackage{pgfplotstable}

\tikzstyle{process} = [rectangle, rounded corners, minimum width=3.5cm, minimum height=1cm, text centered, draw=black, fill=gray!10]
\tikzstyle{arrow}   = [thick, ->, >=Stealth]

\usepackage[numbers,sort&compress]{natbib}
\usepackage[hidelinks]{hyperref}
\usepackage{doi}

\title{MicroDetect-Net (MDN): Leveraging Deep Learning to Detect Microplastics in Clam Blood, a Step Towards Human Blood Analysis}

\author{ \href{https://orcid.org/0009-0000-0660-0849}{\includegraphics[scale=0.06]{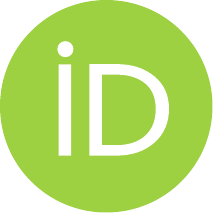}\hspace{1mm}Riju~Marwah}\thanks{Corresponding author.} \\
	Computer Science Engineering\\
	Guru Gobind Singh Indraprastha University\\
	110078, Delhi, India \\
	\texttt{marwah.riju@gmail.com} \\
	\And
	\href{https://orcid.org/0009-0008-9692-0034}{\includegraphics[scale=0.06]{orcid.pdf}\hspace{1mm}Riya~Arora} \\
	Computer Science Technology\\
	Maharaja Agrasen Institute of Technology\\
	110086, Delhi, India \\
	\texttt{riyaarora200311@gmail.com} \\
	\And
	\href{https://orcid.org/0000-0002-7418-7571}{\includegraphics[scale=0.06]{orcid.pdf}\hspace{1mm}Dr. Navneet~Yadav} \\
	Computer Science Technology\\
	Maharaja Agrasen Institute of Technology\\
	110086, Delhi, India \\
	\texttt{navneetyadavdr@gmail.com} \\
	\And
	\href{https://orcid.org/0009-0005-6006-2929}{\includegraphics[scale=0.06]{orcid.pdf}\hspace{1mm}Himank~Arora} \\
	Independent Researcher \\
	\texttt{arorahimank1234@gmail.com} \\
}

\renewcommand{\shorttitle}{Innovative Computation in Biomedical Imaging, ICICC-2025}

\hypersetup{
  pdftitle={MicroDetect-Net (MDN): Leveraging Deep Learning to Detect Microplastics in Clam Blood},
  pdfsubject={q-bio.NC, q-bio.QM, Environmental Science},
  pdfauthor={Riju Marwah, Riya Arora, Dr. Navneet Yadav, Himank Arora},
  pdfkeywords={Deep Learning, Microplastics, Clam Blood, Human Blood Analysis},
}

\date{}

\begin{document}
\maketitle

\begin{abstract}
With the prevalence of plastics exceeding over 368 million tons yearly, microplastic (MP) pollution has grown to an extent where air, water, soil, and living organisms have all tested positive for microplastic presence. These particles, which are less than \textless{}5\,mm in size, are no less harmful to humans than to the environment. Toxicity research on microplastics has shown that exposure may cause liver infection, intestinal injuries, and gut flora imbalance, leading to numerous potential health hazards. 
This paper presents a new model, MicroDetect-Net (MDN), which applies fluorescence microscopy with Nile Red dye staining and deep learning to scan blood samples for microplastics. Although clam blood has certain limitations in replicating real human blood, this study opens avenues for applying the approach to human samples, which are more consistent for preliminary data collection.
The MDN model integrates dataset preparation, fluorescence imaging, and segmentation using a convolutional neural network (CNN) to localize and count microplastic fragments. The combination of CNN and Nile Red dye for segmentation produced strong image detection and accuracy. MDN was evaluated on a dataset of 276 Nile Red-stained fluorescent blood images and achieved an accuracy of \textbf{92\%}. Robust performance was observed with an \textbf{Intersection over Union (IoU)} of \textbf{87.4\%}, \textbf{F1-score} of \textbf{92.1\%}, \textbf{Precision} of \textbf{90.6\%}, and \textbf{Recall} of \textbf{93.7\%}. These metrics demonstrate MDN’s effectiveness in the detection of microplastics.
\end{abstract}


\section{Introduction}
Plastic production worldwide has been rising at an alarming rate, having surpassed 368 million tons per year, with the recent COVID-19 pandemic having highlighted our almost debilitating dependence on plastics. The World Health Organization (WHO) estimates that 89 million medical masks will be required monthly for COVID-19 response. Disposable plastic packaging and personal protective equipment that are disposed of inappropriately, threaten the environment. A critical component of a plastic contamination chain of events is microplastics. These tiny plastic particles are typically less than 5 mm in diameter and either originate from the degradation of bulk plastics or are manufactured directly on a micro- or nanoscale. MP has been increasingly found in marine and freshwater ecosystems. They are also found in air and soil. MP is likewise ubiquitous in foods directly associated with human life, like drinking water, vegetables, fruit, and seafood. This study will detect microplastics in human blood. Clam blood samples have been taken to implement the required research. 

Conventional methods, such as Raman Spectroscopy and Mass Spectrometry, have significant sensitivity but rely heavily on instruments and expertise and are unsuitable for high-throughput use. A novel model, MicroDetect-Net (MDN), detects microplastics in blood. The proposed model works with fluorescence-based microscopy along with Nile Red staining. Our model presents a straightforward, low-cost alternative to visualize and map microplastics in biological tissues using deep learning techniques.

\section{Literature Review}
The following papers have been used for a survey about MP.

\begin{table}[H]
    \caption{Survey Papers on Microplastic Detection in Humans}
    \centering
    \begin{tabular}{P{2.8cm} P{2.8cm} P{5.2cm} P{4.5cm}}
        \toprule
        \textbf{Author} & \textbf{Title / Algorithm} & \textbf{Description} & \textbf{Research Gap} \\
        \midrule
        Kutralam-Muniasamy et al., 2023 & Not specified & Explores progress, problems, and prospects in diagnosing microplastics in humans. & Lacks quantitative methods for microplastic detection in different human samples. \\
        Leonard et al., 2024 & \textmu{}FTIR (micro-Fourier Transform Infrared Spectroscopy) & Characterizes polymer types and concentrations in human blood using advanced \textmu{}FTIR techniques. & Limited data on long-term health implications of microplastic exposure. \\
        Çobanoğlu et al., 2021 & Cytogenetic techniques & Examines genotoxic and cytotoxic effects of polyethylene microplastics on human lymphocytes. & Focuses only on polyethylene microplastics; other types remain unexplored. \\
        Leslie et al., 2022 & High-resolution mass spectrometry & Quantifies and identifies plastic particles in human blood using advanced mass spectrometry. & Does not address potential pathways of microplastic exposure. \\
        Mintoo \& Kousar, 2024 & Not specified & Summarizes existing reports on microplastic presence in human blood. & Does not provide experimental data or analysis methods. \\
        Mastad, 2022 & Not specified & Analyzes evidence from feces and blood on microplastic impacts in humans. & Lacks advanced detection techniques for blood analysis. \\
        \bottomrule
    \end{tabular}
    \label{tab:survey_papers}
\end{table}

\section{Methodology}
This study presents a new model that uses deep learning to locate and segment microplastic fragments in blood samples collected from clams, which could be extended to examine human blood. The use of fluorescence microscopy combined with staining with the Nile Red shows the ability of the model to detect MP. Segmentation and detection of microplastic particles were implemented using Python-based CNN models. This MDN model includes blood sample preparation, fluorescence imaging, and detection of MP particles, as shown in Figure 1. It utilizes clam blood as a proxy, but it suggests a great possibility in this research to transfer clam blood findings to future studies with actual human blood.

\begin{figure}[h]
    \centering
    \includegraphics[width=0.8\textwidth]{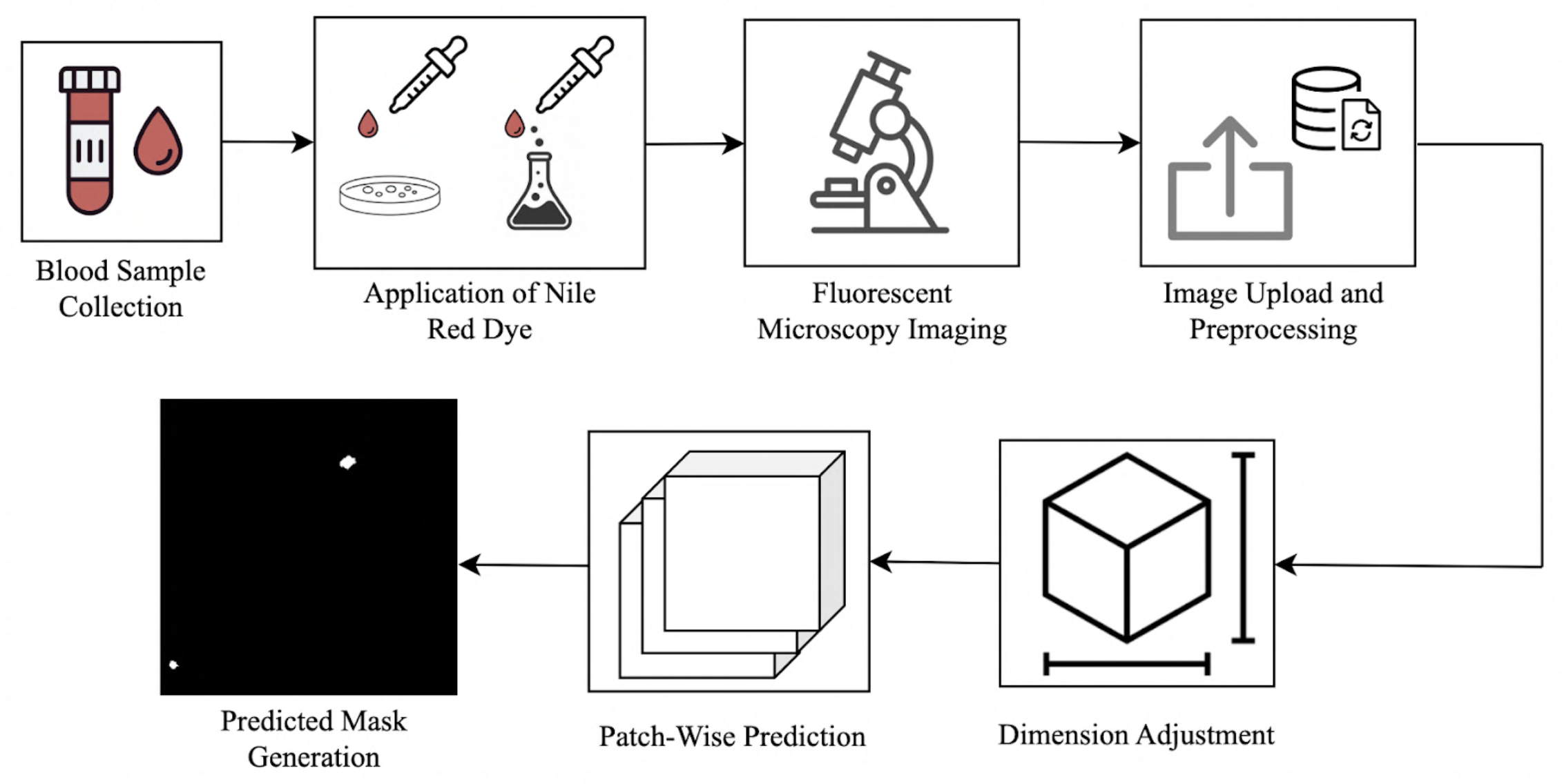}
    \caption{Working of the proposed MDN Model}
    \label{fig:mdn_model}
\end{figure}

\subsection{Nile Red Staining and Fluorescence Imaging}
Nile Red (C\textsubscript{20}H\textsubscript{18}N\textsubscript{2}O\textsubscript{2}) is a synthetic dye initially derived from Nile Blue, which is generally used in the biomedical field to enhance the visibility of the desired elements in the given sample, especially for cells and tissues. Nile red dye is an effective lipophilic stain due to its hydrophobic nature, enabling it to repel water. This property is advantageous for accurately identifying potential microplastic particles, as most microplastics are hydrophobic. The dye binds efficiently to hydrophobic materials like polymers and natural fats, enhancing its ability to differentiate microplastics from other substances. Considering all these factors, Nile Red dye was chosen to stain the collected blood sample. This made MP and other hydrophobic materials visible as fluorescent particles within the blood matrix. In addition, the solvatochromic properties of this chemical allowed plastic classification based on polarity, which increased its usefulness for this work. 

This dye tends to give better results than the other staining chemicals currently in the market. Other dyes like BODIPY dye and Coumarin-Based Dye do mark undesirable blood components whereas Nile Red Dye gives precise and accurate marking of the potential polymer and harmful particles in the human blood which further enhances the accuracy of the model and additionally offers Polarity Sensitivity. 

For Nile Red dye-stained, spiked datasets, distinct fluorescence signals were generated from High-density polyethylene (HDPE) and polyethylene terephthalate (PET) particles. While this study was conducted using clam blood as a medium, the fluorescence spectra of Nile Red can equally be used in human blood, thus paving the way for future use in humans.

\subsection{Dataset Preparation}
The dataset consists of Nile Dye stained fluorescence microscopy images made of both spiked dataset and real MP dataset. The spiked dataset offered controlled circumstances for model training and the real MP dataset populated variability similar to a real-world setting. These datasets enabled the model to acquire knowledge and efficiently generalize the microplastic segmentation task in different environments.

\subsubsection{Spiked Dataset}
The spiked dataset was created by staining pre-microplastic particles using existing types and dimensions of high-density polyethylene (HDPE, 500 $\mu$m) and polyethylene terephthalate (PET, 120 $\mu$m) with Nile Red dye. Fluorescence images were captured and mapped with binary masks manually by three experts. With this dataset, authors ensure conditions were controlled for training and evaluation of the model.

\begin{figure}[h]
    \centering
    \includegraphics[width=0.4\textwidth]{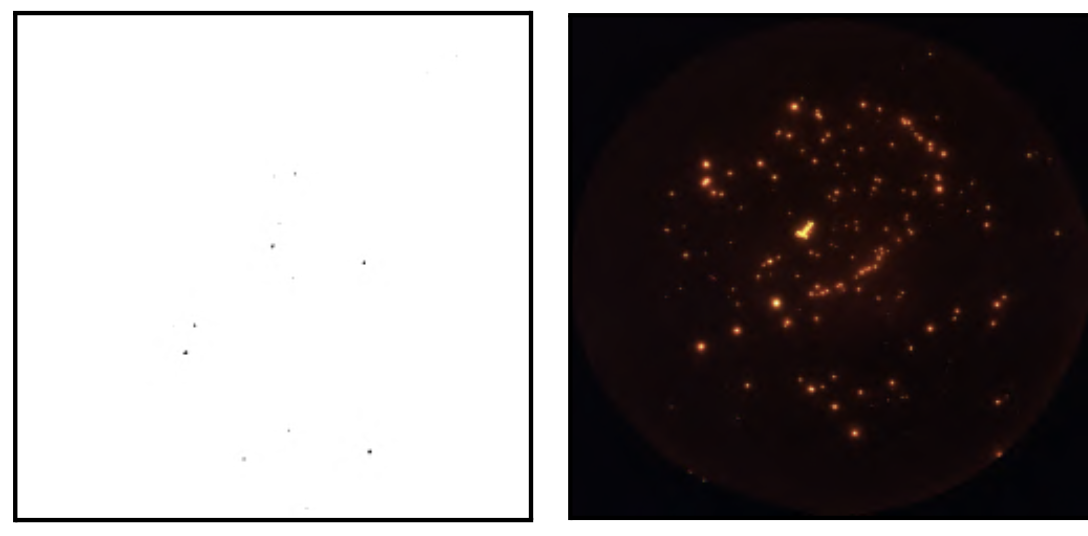}
    \caption{Spiked MP Image}
    \label{fig:spiked_mp}
\end{figure}

\subsubsection{Real Microplastic Dataset}
The MP dataset consisted of microplastic particles isolated from Manila clams, which were real contamination. The isolated particles were stained with Nile Red dye, and numerous fluorescent microscopy images with appropriate binary masks were produced. These images caused a variety of particle sizes, shapes, and densities, which proved to be a good testing measure for the model.

The two datasets were then divided into training (80\%) and testing (20\%) subsets to obtain
balance representation during evaluation. This structure made it easier for the model to
move from datasets that entail controlled spiking to complex and diverse real samples.

\begin{figure}[h]
    \centering
    \includegraphics[width=0.4\textwidth]{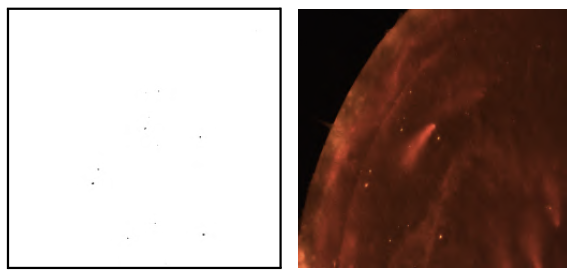}
    \caption{Real MP Image}
    \label{fig:real_mp}
\end{figure}

\subsection{Model Development and Training}
A deep learning-based system is proposed to detect microplastics in blood samples. The system utilizes Convolutional Neural Network (CNN) architecture, implemented using PyTorch, for predicting and segmenting microplastic particles from fluorescence-stained blood samples. The model employs a UNet-based framework, which effectively combines with the CNN model to ensure necessary feature extraction and prediction of the microplastic particles in the blood sample. MDN employs a two-step approach for microplastic particle detection. The first step involves feature extraction using a UNet architecture, designed to effectively capture spatial and contextual features from the input data. In the second step, a CNN is employed to predict and classify microplastic particles based on the extracted features. This integrated approach ensures accurate detection and robust performance in identifying microplastic particles.

\subsubsection{UNet Model Architecture}
The utilized segmentation model used the UNet architecture since it is considered adequate for analyzing biomedical images. The architecture is built as a standard encoder-decoder with skip connections to retain crucial spatial information during feature extraction and reconstruction. It utilizes ResNet-101 as the encoder, which consists of many convolutional and max-pooling layers that shrink in spatial dimensions so that spatial and contextual features are captured. The decoder upsamples the encoded features using transposed convolutions and then fuses them with high-resolution features from the encoder across skip-residual connections, enabling accurate localizations. This UNet model with ResNet-101 extracts all the necessary features for passing to the CNN model for performing the patch-wise MP particle prediction.

\subsubsection{CNN Model}
After receiving the necessary spectacle features from the stained blood sample fluorescence image using the U-Net model, the features are passed to a CNN model. This CNN is trained on 51,513,233 features obtained during the training process from the training dataset by applying numerous convolutional layers optimized to distinguish the microplastic particles. This model classifies the microplastic particles based on features extracted through the U-Net model. In the end, the model provides the output of a microplastic mask that shows the areas where the microplastic target is present. This binary-ended output can measure the extent of microplastic contamination in the images.

This model is trained on fluorescence-stained clam blood sample images, which have been resized to an input size of 256 x 256 x 3 to make it compatible with the convolutional neural network's expectations. This makes microplastic segmentation accurate. The model was trained on 20 epochs using a batch size of 35. We used Stochastic Gradient Descent as an optimizer, and it was successful in achieving its efficiency.

\subsection{Implementation Pipeline}
The process starts with uploading the stained blood sample fluorescence image in the software. Once the image is uploaded it goes through a series of pre-processing steps within the context of the UNet model.

a) Images were normalized to ensure a degree of consistency across the data-pool, by adjusting the intensity levels of the images to a congruent level.

b) By transforming them to 256$\times$256 pixels, we standardized the images and then input them into the U-Net model

c) The image is prepared for patch-based prediction by adjusting it utilizing the following steps. It ensures that the image is perfectly divisible into fixed-size (P) patches of 256$\times$256 pixels. If either of the dimensions is not a perfect multiple of the patch size, leftover portions will remain that will not fit into patches. To counter this, add extra pixels to the required dimension(s) to align them with the closest multiples of the patch size. The final image achieved will be perfectly divided into patches. The following equation can write this process:

\begin{align}
    A_W &= P - (W \mod P) \\
    A_H &= P - (H \mod P)
\end{align}

where $A_W$ is the adjusted width, and $A_H$ is the adjusted height.

\begin{figure}[h]
    \centering
    \includegraphics[width=0.6\textwidth]{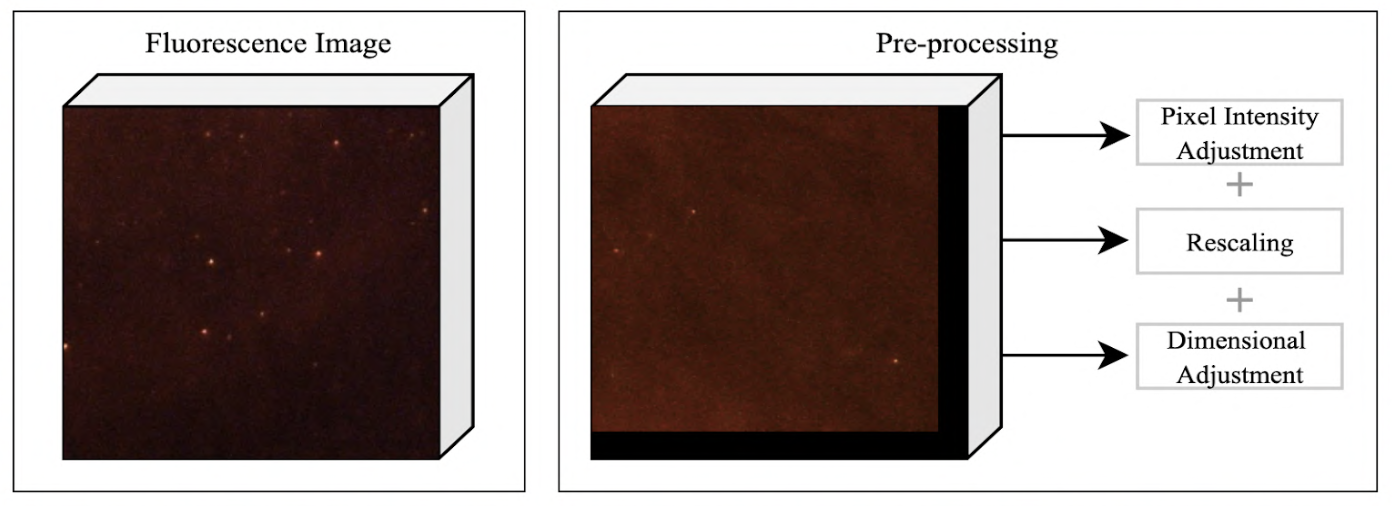}
    \caption{Fluorescence images pre-processing}
    \label{fig:fl_pre_processing}
\end{figure}

After pre-processing, the fluorescence image is input into the U-Net model for patch-wise feature extraction. The extracted features are subsequently aggregated and passed to the CNN model, which generates the final binary mask. This mask distinctly highlights microplastic particles, marking them in white for precise identification and segmentation.

\begin{figure}[htbp]
\centering
\scalebox{0.75}{
\begin{tikzpicture}[node distance=1.3cm and 2cm]
\node (input) [process] {Input Image};
\node (features) [process, below=of input] {Feature Extraction};
\node (downsampling) [process, below=of features] {Downsampling};
\node (bottleneck) [process, below=of downsampling] {Bottleneck};
\node (attention) [process, right=of bottleneck] {Attention};
\node (upsampling) [process, right=of attention] {Upsampling};
\node (skip) [process, above=of upsampling] {Skip Connections};
\node (prediction) [process, above=of skip] {Trained Model Prediction};
\node (mask) [process, above=of prediction] {Detected Microplastic mask};
\draw [arrow] (input) -- (features);
\draw [arrow] (features) -- (downsampling);
\draw [arrow] (downsampling) -- (bottleneck);
\draw [arrow] (bottleneck) -- (attention);
\draw [arrow] (attention) -- (upsampling);
\draw [arrow] (upsampling) -- (skip);
\draw [arrow] (skip) -- (prediction);
\draw [arrow] (prediction) -- (mask);
\end{tikzpicture}
}
\caption{Technical pipeline of the MDN model.}
\label{fig:microplastic_pipeline}
\end{figure}
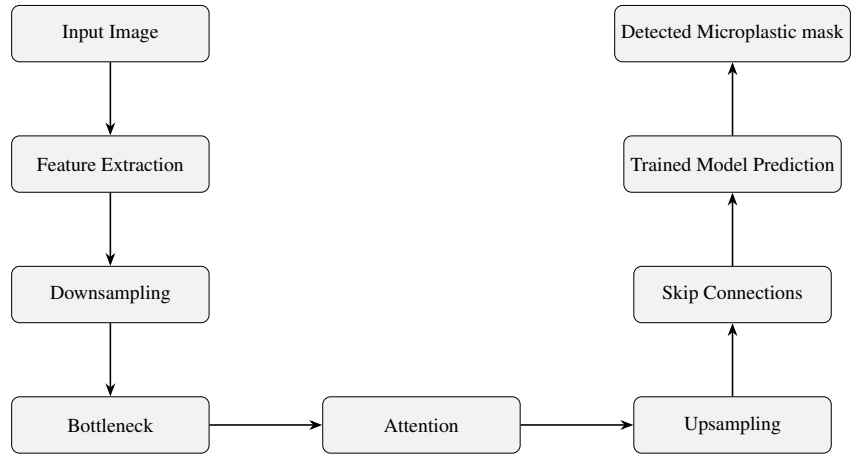

\section{Results}
Our proposed CNN-based model was evaluated on a dataset of 276 images of Nile Red dye-stained fluorescent blood samples, and its ability to detect microplastic particles in blood was assessed. The output is predicted using the parameters Intersection over Union(IoU), F1-Score, Precision, and Recall, as shown in Table~\ref{tab:results}. 

\begin{table}[htbp]
    \caption{Performance metrics of the MDN model}
    \centering
    \begin{tabular}{lc}
        \toprule
        \textbf{Parameters} & \textbf{Percentage} \\
        \midrule
        Intersection over Union (IoU) & 87.4\% \\
        F1-Score                      & 92.1\% \\
        Precision                     & 90.6\% \\
        Recall                        & 93.7\% \\
        \bottomrule
    \end{tabular}
    \label{tab:results}
\end{table}

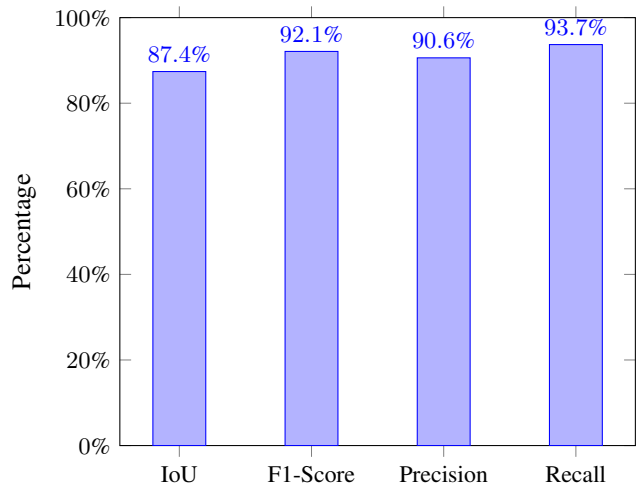
\begin{figure}[H]
    \centering
    \begin{tikzpicture}
        \begin{axis}[
            ybar,
            bar width=20pt,
            ymin=0, ymax=100,
            ylabel={Percentage},
            symbolic x coords={IoU, F1-Score, Precision, Recall},
            xtick=data,
            nodes near coords,
            nodes near coords style={font=\small},
            nodes near coords={\pgfmathprintnumber[fixed,precision=1]{\pgfplotspointmeta}\%},
            tick label style={font=\footnotesize},   
            yticklabel={\pgfmathprintnumber[fixed,precision=0]{\tick}\%},
            enlarge x limits=0.15
        ]
        \addplot coordinates {(IoU,87.4) (F1-Score,92.1) (Precision,90.6) (Recall,93.7)};
        \end{axis}
    \end{tikzpicture}
    \caption{Performance metrics of the model}
    \label{fig:model_metrics}
\end{figure}

\section{Discussion}
MicroDetect-Net was evaluated on a dataset of 276 Nile Red-stained fluorescent blood images and achieved an accuracy of 92\%. Robust performance, with an Intersection over Union (IoU) of 87.4\%, F1-score of 92.1\%, Precision of 90.6\%, and Recall of 93.7\%, was observed. These metrics validate MDN’s effectiveness in detecting microplastics.

Clam blood was used in this study in lieu of human blood, as differences between the human and clam medium were acknowledged. For the time being, clam blood was found to be accessible and consistent for preliminary experiments and development. We conclude that localization of Nile Red dye staining and CNN-based segmentation of microplastic particles could also be performed on human blood samples with little modification. Future development upon our research could see this technique being used to detect microplastic contamination in human blood samples in a clinical setting.

Utilizing clam blood as a proxy can have limitations affecting real-world applications. The physiological differences between human and clam blood need to be considered. Red Blood Cells (RBCs), White Blood Cells (WBCs), and platelets are abundant in human blood, while clam blood lacks them. Their presence in human blood can interfere with fluorescence microscopy, potentially obscuring microplastic signals and causing light scattering and absorption. This could reduce detection accuracy.

Moreover, clam blood immunity relies on hemocytes, whereas human blood contains a complex system of WBC subtypes. These immune cells might bind to microplastics, masking them during the detection process. Additionally, human blood contains high levels of lipid content (e.g., HDL, LDL) and complex plasma proteins. These may interact with Nile Red dye, affecting its staining efficiency.

Despite these limitations, mitigating them is possible through the following approaches. Begin validation on human blood to test and refine the model's generalizability. Use an incremental transition approach: begin with clam blood for preliminary research, and gradually integrate human samples. Add human-specific components (e.g., plasma proteins) to clam blood to better approximate real human blood conditions. Expand the dataset using real human blood samples to improve accuracy and reliability.

\section{Conclusion}
Through this work, we successfully built and trained a deep-learning model to detect microplastic contamination in fluorescent microscopy images of Red Nile-dyed blood. Our models automated detection ability makes it a promising prospect for future use in the domain of medical diagnostics, providing a way to quantify the concentration of microplastics in human bloodstream. The methodology presents a great potential for developing diagnostic tools that, in the future, could detect microplastic particles within human blood samples, catering to the existing deficit in research. 

Future opportunities exist for scaling the dataset, refining the model architecture, and exploring deployment strategies to elevate accessibility further, enabling broader adoption and increased impact. Exploration of real-time deployment strategies like integration with IoT devices could enable more widespread use. We may also categorize the various microplastics based on their dimensions. Based on their ferret diameter and dimensions, future research may categorize microplastics by dimensions and assess health impacts accordingly.

Despite the study's achievements, limitations exist, such as the limited dataset size of the clam blood samples used as proxies. Collaboration in this field with an existing deficit in large-scale data collection will certainly impact the application of this study. Given these advancements, the proposed model could greatly contribute to mitigating the dangers associated with microplastic particle presence in various living beings.

\nocite{*}

\bibliography{references}

\end{document}